\documentclass[10pt,twocolumn,letterpaper]{article}

\usepackage[pagenumbers]{cvpr} 

\usepackage{graphicx}
\usepackage{amsmath}
\usepackage{amssymb}
\usepackage{booktabs}
\usepackage[pagebackref,breaklinks,colorlinks]{hyperref}

\usepackage[capitalize]{cleveref}
\crefname{section}{Sec.}{Secs.}
\Crefname{section}{Section}{Sections}
\Crefname{table}{Table}{Tables}
\crefname{table}{Tab.}{Tabs.}

\usepackage[shortlabels]{enumitem}
\usepackage[accsupp]{axessibility}  

\makeatletter
\@namedef{ver@everyshi.sty}{}
\makeatother
\usepackage{tikz}

\usepackage{orcidlink}

\def\cvprPaperID{4}
\def\confName{CVSports}
\def\confYear{2023}

\begin{document}

\title{Monocular 3D Human Pose Estimation for Sports Broadcasts \\using Partial Sports Field Registration}


\author{Tobias Baumgartner~\orcidlink{0000-0003-1194-3429}, Stefanie Klatt~\orcidlink{0000-0002-2477-8699}
\vspace{5pt}
\\
{Institute for Exercise Training and Sport Informatics}\\
{German Sport University, Cologne}\\
{\tt\small t.baumgartner@dshs-koeln.de}
}
\maketitle

\begin{abstract}
The filming of sporting events projects and flattens the movement of athletes in the world onto a 2D broadcast image. The pixel locations of joints in these images can be detected with high validity. Recovering the actual 3D movement of the limbs (kinematics) of the athletes requires lifting these 2D pixel locations back into a third dimension, implying a certain scene geometry. The well-known line markings of sports fields allow for the calibration of the camera and for determining the actual geometry of the scene. 
Close-up shots of athletes are required to extract detailed kinematics, which in turn obfuscates the pertinent field markers for camera calibration. We suggest {\normalfont partial} sports field registration, which determines a set of scene-consistent camera calibrations up to a single degree of freedom.
Through joint optimization of 3D pose estimation and camera calibration, we demonstrate the successful extraction of 3D running kinematics on a 400m track.
In this work, we combine advances in 2D human pose estimation and camera calibration via partial sports field registration to demonstrate an avenue for collecting valid large-scale kinematic datasets. We generate a synthetic dataset of more than 10k images in Unreal Engine 5 with different viewpoints, running styles, and body types, to show the limitations of existing monocular 3D HPE methods. 
Synthetic data and code are available at \url{https://github.com/tobibaum/PartialSportsFieldReg_3DHPE}.
\end{abstract}

\section{Introduction}
\label{sec:intro}

\begin{figure*}[t!]
    \centering
    \includegraphics[width=\textwidth]{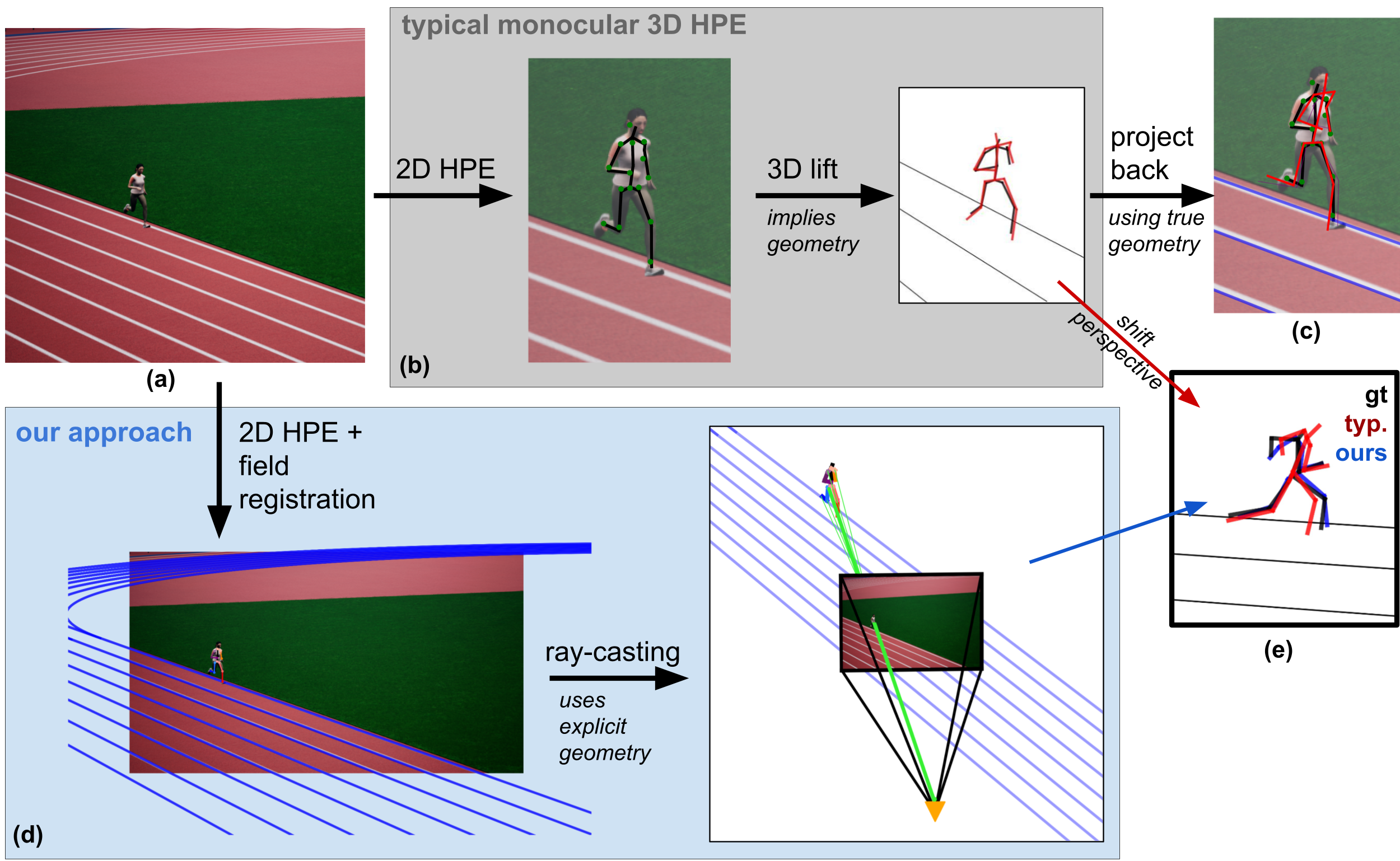}
    \caption{Comparison between typical monocular 3D HPE and our approach. (a) Rendered running scene with known camera calibration and 3D joint locations. (b) In a typical monocular 3D HPE pipeline, the pixel locations of the joints are determined and then lifted into 3D, which implies some geometry. (c) Placing the estimated 3D pose in the actual scene and rendering the projection leads to a significant deviation from the detected 2D pose.
    (d) In our approach, we take input (a) and extract the camera parameters (via sports field registration) and use the detected 3D geometry to cast rays into the scene. (e) The comparison between ground truth (black), a state-of-the-art method (red), and our approach (blue) shows the differences when viewed from an out-of-camera-plane perspective.}
    \label{fig:fig1}
\end{figure*}

2D Human pose estimation (HPE) has reached a level of maturity where it is highly accurate and widely available as models or APIs for developers to integrate into their apps~\cite{openpose,lugaresi2019mediapipe}. Consumer applications in the sports-tech domain use 3D human pose estimation to monitor the user and provide feedback for the safe and successful execution of movements. While pose estimation methods exceed in providing the locations of joints, human motion requires knowledge about sequences of locations and displacements and their interaction. While many biomechanical models of human motion exist~\cite{ackermann2010optimality,faucris.279321780,song2021deep,millard2013flexing}, there is no unified approach towards understanding and facilitating handling human movement on a large scale yet.
The current rise and mainstream success of large language models (\eg ChatGPT~\cite{radford2018chatgpt}, LLaMA~\cite{touvron2023llama}) demonstrates the emergent potential of foundational machine learning models. In the recently pre-published MotionBERT~\cite{motionbert2022}, the authors propose a unified model that encodes motion from different source datasets and is trained to solve a variety of tasks, thereby resulting in a shared representation of motion to be used in downstream applications. Scaling such methods and building a unified representation of human motion, requires large amounts of well-structured and valid kinematic recordings, \eg, from sports broadcasts.
Collecting the required scale of data will only ever be feasible from video, ideally from already existing footage. Sports broadcasts are a promising source of data, and the challenge becomes how to accurately extract kinematic information from these data.

To develop scalable methods that can accurately extract kinematically valid data, benchmarks are necessary to evaluate the progress of 3D HPE methods. However, ground truth data is expensive and not readily available, and it is challenging to collect datasets for 3D human pose estimation. This creates a circular dependency, as we need data to test the validity of the methods, but we can only collect the necessary data at scale if gold-standard methods exist. To break this circular dependency, we propose evaluating current methods on a synthetic benchmark. 

While monocular 3D HPE is widely used and well-studied~\cite{rie2021,li2022stridedtrans,DBLP:journals/corr/abs-2107-07797, Xu_2020_CVPRdeepkinematics,li2022stridedtrans}, it has recently been shown to not be sufficiently precise for very specific sports use cases, e.g., analyzing intra-athlete variations as signs of fatigue~\cite{baumgartner2023}. Since the variations within athletes can be very small, the error in the 3D human pose estimation process needs to be equally small to detect relevant differences.

In our work, we focus on the specific domain of middle-distance running. Running is a fundamental motion and an integral part of many sports. Additionally, it readily lends itself to this investigation, as the motion itself is very cyclical and does not vary much in professional athletes over the course of a short video sequence, which helps to reveal problems in current approaches. 
There is no ground truth 3D human pose estimation data available for typical middle-distance running scenarios on the track.

We render a synthetic dataset similar to typical track \& field recordings using Unreal Engine 5 (\cf \cref{fig:fig1,fig:perspective}), to generate a ground truth dataset for demonstration purposes.
In our experiments, we will demonstrate that a state-of-the-art monocular 3D HPE method~\cite{Sarandi2023dozens} does not generalize well to this scenario. In \cref{fig:fig1}, we illustrate the rationale for why this is. Typical 3D HPE methods perform 2D HPE and then lift the 2D pixel locations into 3D by a learned function. This lifting step implies some geometry (even if not explicitly modeled), which is indicative of the inductive bias introduced during the training of the method. When placing the estimated 3D skeleton into a known scene, and rendering its projection, we see in \cref{fig:fig1}(c) that the re-projection does not align with the originally used 2D skeleton. It follows, that the implied geometry of the 3D HPE method must have been different from the known, correct geometry.

Learning about the geometry in a scene (\ie camera calibration) is well-studied in the realm of sports, and achieved via sports field registration~\cite{chen2019sports,TVCalib22,chu2022sports}. Since many sports are played on a field or track with pre-defined dimensions, camera calibration can be achieved by fitting a virtual camera to the visible lines in broadcast footage. Sports field registration is usually performed on wide camera shots that show multiple non-co-linear sections of the field, like corners or end-zone/penalty area markings. On the other hand, kinematic data is best extracted from close-up viewpoints that highlight single athletes. Existing methods for sport-field registration are, therefore, not necessarily applicable to our use case.
The typical broadcast camera setup in track~\&~field events are stationary cameras that pan, tilt, and zoom to follow the athletes. During these camera motions, only small portions of the track around the athlete are in frame (\cf \cref{fig:fig1}(a)), and full sports field registration is infeasible.

We use a constructive method to find a complete set of valid camera calibrations that are consistent with just a single vanishing point derived from lanes in the image~\cite{baumgartner2023}. This set of camera calibrations has one remaining degree of freedom, which makes our approach a \textit{partial} sports field registration method. Using consistency constraints about a sequence of frames, we can jointly derive complete field registration. As illustrated in \cref{fig:fig1}(d), we use explicit knowledge about the 3D geometry of the current frame to cast a ray through the image and lift a 2D skeleton into 3D. The resulting 3D skeleton is, by construction, consistent with its 2D re-projection (as opposed to typical methods, \cf \cref{fig:fig1}(c)), and yields a closer 3D estimation in our scenario than previous methods (\cf \cref{fig:fig1}(e)).

Overall, we make the following contributions:
\begin{enumerate}[nosep, wide, labelwidth=0pt, labelindent=0pt]
    \item We generate a novel synthetic running dataset, based on Unreal Engine 5, Mixamo, and Metahumans. We provide video renderings with ground truth annotations for 2D for joint pixel locations, absolute 3D world joint coordinates, and camera calibrations.
    \item We demonstrate that state-of-the-art monocular HPE methods work well on this data in the case of 2D annotations, but invoke considerable errors in the 3D case.
    \item We develop a novel 2D to 3D lifting method that uses 2D annotations together with the overall scene geometry to ray-cast absolute joint locations into the scene. This allows for improving the 3D estimation of existing methods. 
\end{enumerate}

\section{Related}
\label{sec:related}


\textbf{Human pose estimation (HPE).}
In the field of 2D HPE, publicly available images can be manually annotated using crowd-sourcing methods to create datasets and benchmarks for evaluation~\cite{coco,mpii}, resulting in state-of-the-art methods that closely approach human-level performance~\cite{openpose,wholebody,xu2022vitpose}. However, for 3D HPE, collecting ground-truth pose data requires a more complex lab setup. The popular \textit{Human3.6m} dataset, for instance, contains 11 actors in 17 scenarios with 4 synchronized cameras and marker-based motion-capture, limiting its scope to close-up shots, despite its large size of 3.6 million images~\cite{h36m_pami}.

A common strategy to tackle monocular 3D HPE is based on video data. If a system can track a 2D skeleton over consecutive frames, it can merge these frames and resolve the joint task of forecasting a 3D skeleton that accounts for all the 2D projections of the individuals in the scene~\cite{rie2021,li2022stridedtrans,DBLP:journals/corr/abs-2107-07797}. Works in this direction aim for consistency by leveraging limb length~\cite{rie2021}, realistic motion~\cite{Xu_2020_CVPRdeepkinematics}, or temporal context~\cite{li2022stridedtrans}.
Gong \etal propose a technique to enhance the limited existing datasets of 2D/3D human pose estimation pairs with additional modifications for viewpoint, posture, or body size to improve the training process~\cite{poseaug}.
Another category of methods aims to infer the 3D information of human poses by concurrently predicting the depth of the current image~\cite{sarandi2021metrabs,DBLP:journals/corr/PavlakosZDD16,DBLP:journals/corr/abs-1711-08229,luvizon2022consensus}. These methods do not necessarily require video data. By anticipating the distance between each object, person, and surface in the frame to the camera, these techniques produce a so-called 2.5D image that can be combined with 2D HPE to infer 3D poses.
Recent approaches handle these two tasks simultaneously, which introduces a bias on the possible configurations of the 3D skeleton. 
In their investigation~\cite{sarandi2021metrabs,Sarandi2023dozens}, Sárándi \etal also incorporate the intrinsic camera parameters to develop a cutting-edge system for monocular 3D HPE in the wild, which they call \textit{MeTRAbs}.

\textbf{Sports field registration.}
Previous works have performed camera calibration on existing broadcast footage for analyzing sporting events by sports field registration. Chen and Little use the known dimensions and line markings of fields in ball sports to generate synthetic data for training a Siamese network~\cite{chen2019sports}. Chu \etal solve this task using sparse keypoint detections~\cite{chu2022sports}. Theiner and Ewerth proposed a single-shot method that calculates a homography to determine the camera calibration~\cite{TVCalib22}. However, in our scenario, where only the distance between parallel lines is known and co-linearity of input data prevents homography calculation, we focus on the subtasks of lane detection and vanishing points. Our approach presents a unique challenge in camera calibration for sports analysis.

\textbf{Rendered data for computer vision.}
There is precedence for using rendered data for creating datasets and benchmarks. UnrealCV~\footnote{\url{https://unrealcv.org/}} is a plugin for Unreal Engine 4 that can be used to create benchmarks for object detection or object pose estimation~\cite{qiu2017unrealcv}. The authors of Human3.6m also provide mixed reality scenarios in which they render 3D models of poses recorded in the lab into real-world scenes~\cite{h36m_pami}. To the best of our knowledge, game engines have not been previously used to render complete scenes for generating ground truth for 3D HPE.

\section{Method}
\label{sec:method}

\begin{figure}[t]
    \centering
    \includegraphics[width=.45\textwidth]{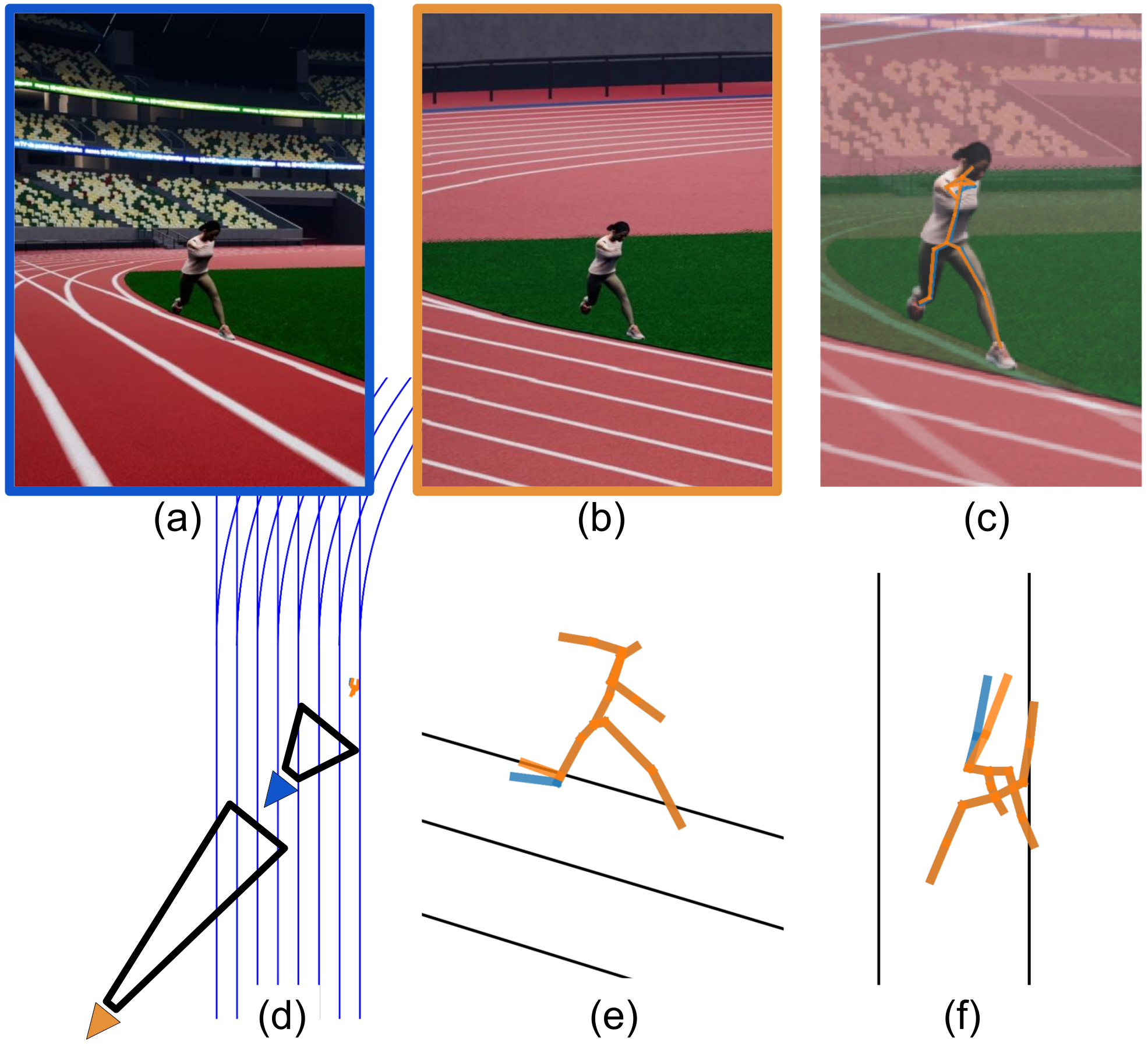}
    \caption{Illustrating the ambiguity between vastly different camera perspectives. (a) Close-up shot of a certain step (blue camera). (b) Highly zoomed shot of a very similar situation (orange camera). The athlete's right foot has been modified to result in the same 2D image of the athlete, as indicated by the alpha-blend in (c). (d) The two cameras are positioned at vastly different distances and have corresponding field-of-views. (e) The 3D limb configuration differs by 7.8$^\circ$, which is more pronounced from the top-down view in (f).}
    \label{fig:perspective}
\end{figure}

In this work, we introduce a novel method for monocular 3D human pose estimation that is specifically designed for the large-scale collection of valid kinematic data from middle-distance running broadcasts.
We demonstrate how injecting knowledge about the 3D geometry of a scene can improve the precision of 3D HPE. 
The process of recording sporting footage flattens information about the 3D world onto a 2D screen. Typical state-of-the-art approaches towards 3D HPE undo this projection by lifting detected 2D joint locations back into a third dimension (\cf \cref{fig:fig1}(b)).
In this process, some information about the 3D setup of the current image is implied. Whether the monocular 3D HPE method uses a certain camera-model~\cite{sarandi2021metrabs} or inter-frame consistency~\cite{rie2021,li2022stridedtrans}, it has to somehow inflate information into a third dimension. The inductive bias from respective methods is learned over the statistics of the underlying training set.
Most of the existing 3D HPE approaches are trained on datasets recorded in a laboratory setting with multiple static cameras, or in the wild with stereo-vision camera setups. In both of these approaches, the recorded subject is roughly within 6-20 meters of the camera. In the domain of sports broadcasting, the distances between the camera and the athletes are significantly larger and tele-zoom lenses with a narrow field-of-view are commonly used. It is to be expected, that off-the-shelf monocular 3D HPE methods will perform worse on these out-of-distribution images. 

A typical remedy to overcome this issue would be to fine-tune existing methods with 2D-3D correspondences in the required camera setting with large camera-athlete distances. This is a prohibitively complex task, as there are vast amounts of possible variations in camera settings, and collecting ground truth either requires setting and precisely calibrating multiple cameras in a stadium or for athletes to wear portable motion capture suits.

To motivate our approach, we first acknowledge, that different camera settings can lead to the exact same projected 2D points, even though the underlying 3D constellation of the athlete's limbs were different. This error in 3D projection is admittedly small but leads to an expected error in measurement that is prohibitively large to use the resulting 3D pose estimation for kinematic investigations. In \cref{fig:perspective} we illustrate a scenario in which the same athlete in (a) and (b) occupies the exact same pixels in two different broadcast shots, which is shown by the blended image in (c). The respective cameras taking these pictures are located as indicated in (d). In order for the 2D images to be the same, we manually rotated the right leg of the athlete, which can be seen in (e) and the top-down view in (f). The difference between the leg in the two images is an angle of 7.8$^\circ$. This constructed scenario demonstrates, that the exact same image pixels could have resulted from significantly different 3D skeletons, based on the exact camera calibration. The 2D HPE alone does not contain enough information to distinguish between these two. The deflated information about the 3D world is irrevocably lost.

In our approach, we use additional clues about the scene to disambiguate between camera calibrations and, thereby, between potential 3D estimations. We claim that 2D pose estimation by itself does not carry enough information about a scene to recover kinematically valid 3D pose estimations.


\subsection{Synthetic ground truth}
Advances in computer graphics lead to publicly available game engines that allow for almost photo-realistic renderings of arbitrary scenes. We use \textit{Unreal Engine 5}\footnote{\url{https://www.unrealengine.com/unreal-engine-5}} together with animations from \textit{Mixamo}\footnote{\url{https://www.mixamo.com/}} (based on marker-based MoCap) and characters generated with \textit{MetaHumans}\footnote{\url{https://metahuman.unrealengine.com/}} to generate a dataset of over 10k images. The characters generated using MetaHumans look realistic enough to be easily recognized as humanoid by current person detection methods and we can perform 2D and 3D HPE on the rendered images.

We render a dataset consisting of 31 sequences with characters varying in height, body composition, limb length, and running style. Each sequence shows an athlete running down the straight of a 400m track from different camera positions. The static camera pans, tilts, and zooms similar to a broadcast recording, but does not necessarily point directly at the athlete. Each sequence is manually constructed and animated. We use custom Unreal Engine blueprints to extract the 3D world location of each joint and the complete camera calibration (location, orientation, field-of-view).
We extract 3D joint locations by using the character's rig, \ie, the exact location of their artificial bones, and map them to the joint definitions from Human3.6m~\cite{h36m_pami}. We project the 3D joint locations into 2D pixel locations using the extracted camera calibrations. Overall, we generated a ground truth dataset with (1) rendered images, (2) 2D pixel locations of joints, (3) 3D world coordinates of joints, and (4) camera calibrations for 10571 frames.

\subsection{Partial sports field registration on 400m track}
As evident by most figures in this paper, we approach the domain of middle-distance running. In broadcast footage on the 400m track, the lane demarcations are clearly visible, even though only a small portion of the overall track is contained in each frame. Since the visible lines are co-linear, it is neither possible to pinpoint their exact location on the track, nor can the camera calibration unequivocally be determined from just these parallel lanes. We use and improve a recently proposed method~\cite{baumgartner2023} to derive a dense set of potential camera calibrations using only the visible parallel lanes.

As elaborated previously, the close-up shots in our targeted scenarios only allow for partial views of a field, or the 400m track of a stadium. In our approach, we locate the track lane markers (\cf \cref{fig:anecdote}, blue) and use them to determine a single vanishing point. Using the method described by XX \etal, we generate a set of candidate camera calibrations $\mathbb{P}$, which each describe a virtual camera $C$, and are consistent with this vanishing point and the visible lanes~\cite{baumgartner2023}. This set is constructed in a way that each $\mathbf{P^C} \in \mathbb{P}$ describes one azimuth of the camera to the scene, in increments of 1$^\circ$. The other components of $\mathbf{P^C}$ (elevation, roll, field-of-view, camera location) are calculated to match the azimuth and be consistent with the visible image:

\begin{enumerate}[nosep, wide, labelwidth=0pt, labelindent=0pt]
    \item Create a lookup-grid of vanishing points when sweeping the azimuth and elevation under fixed field-of-view. 
    \item Determine the visible track lanes (\cf \cref{fig:anecdote}, blue) and their vanishing point $v_0$. 
    \item For each azimuth in the lookup-grid, create a virtual camera $C$ and determine the elevation and roll that result in a vanishing point in the correct \textit{direction}. This results in a rotation matrix $R^C$ for each azimuth.
    \item Alter the distance of the vanishing point to the scene by adapting the field-of-view. We now have a camera $C$ with a partial camera calibration ($R^C$, $K^C$) that results in the same vanishing point as the scene.
    \item Place each of the previous cameras $C$ in a simulated scene and shift it, such that the virtual track markings correspond to the visible lanes. The required transformation describes a translation~$t^C$.
    \item Combing the above steps, compute the overall projection matrix $P^C=K^C\cdot(R^C|t^C)$.
\end{enumerate}

\begin{figure}[t]
    \centering
    \includegraphics[width=.48\textwidth]{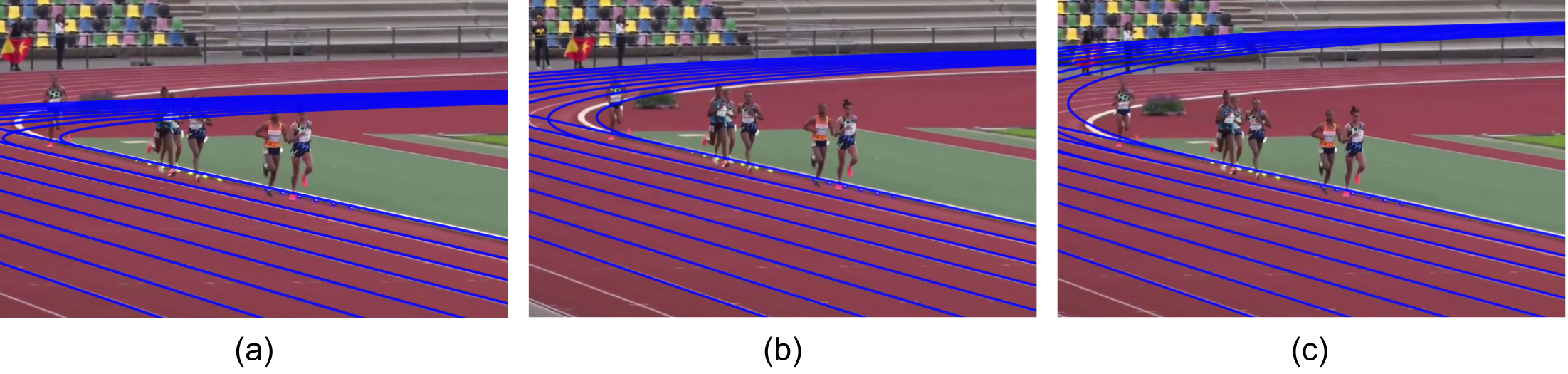}
    \caption{Different exemplar plausible camera calibrations for the same scene. Each potential calibration is consistent with the lane demarcations but differs in camera location, field-of-view, and angle to the scene. (a) Low camera height. (b) Approximately correct height. (c) High camera angle.}
    \label{fig:calibration}
\end{figure}

\cref{fig:calibration} shows three different valid calibrations for a single frame. We computed a collection of camera calibrations with 1$^\circ$ difference in azimuth, but can freely interpolate between these to allow for an arbitrary, continuous selection of camera calibration, which can be interpreted as a slider that determines the distance of the second vanishing point. We hence have arrived at a representation with a single degree of freedom for each of the frames.

From this collection of potential calibrations $\mathbb{P}_f$ for each frame, we now pick a virtual camera $C_f$ for each frame, that results in a consistent description of the entire sequence. 
We know that the camera is static, therefore, the correct subset of calibrations for each frame share the same height and orthogonal distance to the track lanes. Our system is designed in a way, such that we can choose any attribute of the camera (field-of-view, location, tilt angle, $\dots$) and adapt its other attributes to match up with the visible lanes. For each sequence, we can pick a single camera height and thereby remove the last degree of freedom from each frame-wise calibration.

\begin{figure*}[t]
    \centering
    \includegraphics[width=\textwidth]{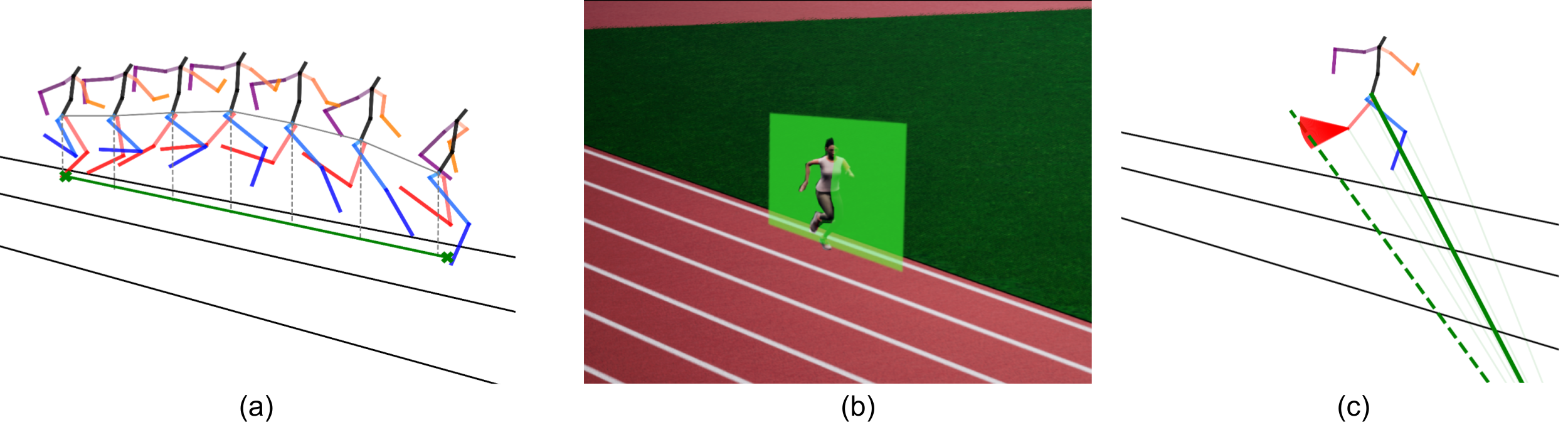}
    \caption{(a) Determining the location of the athlete in the scene. Connections to the ground are marked with green crosses. (b) Sagittal plane of the athlete, even when they do not currently touch the ground. (c) Casting a ray into the scene and intersecting it with a sphere of limb-length radius around the parent joint.}
    \label{fig:method}
\end{figure*}

\subsection{Monocular 3D HPE from sports field geometry}
\label{sec:monoc3dhpe}
Assume we have now picked the correct camera calibration for a frame and know the exact 3D geometry of the scene and how it lines up with the visible image. We can now use this 2D-3D correspondence to cast rays into the scene and determine the exact joint locations of an athlete.

\textbf{Derive sagittal plane.} Whenever the athlete touches the ground, we can determine the exact location of their foot by casting a ray through its pixel location into the scene, intersecting it with the ground plane. In between two connections with the ground, the flight path of the athlete does not change (unless hit by another athlete, which we rule out for now).
We can therefore interpolate the ground-projection of the athletes' location in intermediate frames (\cf \cref{fig:method}(a)). We use this location to determine the athletes' sagittal plane (\cf \cref{fig:method}(b)). 

\textbf{Cast pelvis into the scene.} Using this sagittal plane of the athlete, we can now determine the absolute location of their pelvis, even when they are not touching the ground by casting a ray into the scene. We find the point of intersection between a ray from the camera center through the pixel location of the pelvis into the sagittal plane of the athlete (\cf~\cref{fig:fig1}(d)). Ray-casting only works, when knowing which plane in the scene to intersect with. Typically, this can be done for feet, as during steps it can be assumed, that they are on the ground. Assuming a certain height for the hip would result in faulty localization, since the height of the hip describes a curve, as exaggeratedly indicated in \cref{fig:method}(a). Determining the sagittal plane of the athlete is, therefore, indispensable, as the distance of the athlete to the camera could not be determined otherwise. 

\textbf{Limb consistency.} 
The length of the athletes' limbs does not change between frames. As a rough approximation for their correct limb length, we chose the average length for each limb over the course of the entire scene as extracted with an off-the-shelf monocular 3D HPE method. Starting from the pelvis, we build up the torso and then both legs and arms by again casting a ray into the scene, intersecting it with a sphere with limb length radius around its parent joint location. In \cref{fig:method}(c), we illustrate this process by the example of an athlete's left foot. Assume we have already placed the left knee (light red line). We now cast a ray through the pixel location of the right ankle (dashed green line) and find a point where its distance to the knee is exactly the previously determined length of the shin. For every joint, there are two different points in space at which the limb-length sphere around the parent joint is intersected. This means, that for a body segment with three limbs (arm = shoulder, elbow, wrist, leg = hip, knee, ankle), there are eight potential configurations that are consistent with the image. These options are pruned down by both range-of-motion constraints and frame-to-frame consistency.

\section{Experiments}
\label{sec:experiments}

We evaluate the above method on our newly generated special-purpose dataset to showcase the gaps in current state-of-the-art approaches and a first baseline to solve some of the initially described problems. For this investigation, we chose a very versatile solution by Sárándi \etal which holds the current state-of-the-art in the "3D Poses in the Wild Challenge" (with additional training data, MPJPE)~\cite{sarandi2021metrabs}. In a recent investigation, the authors showcase the slight differences in joint definitions between pose estimation datasets~\cite{Sarandi2023dozens}. We analyze and evaluate all of their implemented joint definitions. The lowest pixel error between 2D joint predictions and our 2D ground truth was achieved using the MS-COCO joint location definitions~\cite{coco}, which resulted in an average pixel error of 4.66$\pm$0.87 pixel over our entire dataset, using their model MeTRAbs-XL.
Throughout our experiments, we use their suggested MeTRAbs model as a base and use our method to improve the pose estimation results.

\subsection{Camera calibration}
Our approach builds on several steps and combines camera calibration with pose estimation. We first evaluate our approach by investigating each step in the method's pipeline (\cf \cref{sec:monoc3dhpe}. We first extract the visible lanes from each frame using a default OpenCV implementation of the linear Hough transform. The endpoints of the detected lanes are on average 21.98$\pm$24.42 pixels off the groundtruth image lanes. This results in an estimation error of the vanishing point of 2.58$\pm$ 1.89\%. Overall, the camera location determined with our partial calibration is 0.85$\pm$0.72~m off the ground truth location, which is 3.97$\pm$3.10\% of the camera's distance to the track lanes. We estimate the field-of-view of an average error of 3.07$\pm$0.88$^\circ$.

\begin{table}[t]
    \centering
    \begin{tabular}{l|ccc}
    \toprule
     Method & Re-projection & 3D Error & Knee angle \\
     & [pixel] & [cm] & [degree] \\
    \midrule
    MeTRAbs &    6.36 (4.08) & 10.33 (1.69) & 20.31 (9.74)\\
    + rotation & 5.01 (2.96) &  7.92 (2.12) & 12.41 (7.94)\\
    \midrule
    Our method & 2.76 & 6.41 (1.65) & 9.91 (9.00) \\
    + context & - & 2.44 (0.58) & 2.87 (3.27)\\
    \bottomrule
    \end{tabular}
    \caption{Result on synthetic ground truth dataset with enhancements. By construction, our method does not have any re-projection error.}
    \label{tab:results}
\end{table}

\subsection{Pose Estimation}
The following experiments are run on our ground truth dataset with 10571 frames consisting of different viewpoints, athlete compositions, and camera movements. We evaluate the various approaches using these metrics:

\textbf{Re-projection.} After estimating a 3D skeleton, we place it into the scene at the exact location of the 3D ground truth (matched pelvis location) and project it into the image using the ground truth camera calibration. The re-projection score describes the offset in pixels from the underlying 2D HPE locations that the 3D skeleton has been computed from. \cref{fig:fig1}(c) shows the offset between the detected 2D joint locations (green dots) and the projected red 3D skeleton. This metric is a proxy for the error in geometry that the 3D lifting process implied~\cite{baumgartner2023}.

\textbf{3D Error.} Next, we measure the difference between estimated and ground truth 3D skeletons in centimeters.

\textbf{Knee angle.} Since the previous euclidean distance itself does not carry much information about the kinematic implications of certain errors, we also measure the specific difference in knee angles, which can directly be interpreted. This measure is crucial for the feasibility of a method to be valid for use in kinematic investigations. In the running literature, effects of angle differences between athletes as low as 3-4$^\circ$ have been reported~\cite{baumgartner2023,kneetreadmill1,moore12beginnerknee,vicon}, and therefore we need to design pose estimation methods that are precise enough to allow for reporting this level of detail.

In \cref{tab:results}, we compare the following methods and improvements:

\textbf{MeTRAbs.} Off-the-shelf \textit{Absolute 3D Human Pose Estimator} and current state-of-the-art~\cite{sarandi2021metrabs}. 

\textbf{+ rotation.} We improve the previous method by including our estimated camera calibration and removing the relative orientation of the athlete to the camera. This improvement already requires information about the 3D geometry of the scene which is only available through our partial sports field registration approach.

\textbf{Our method.} Using our estimated camera calibration, together with pose estimation by ray-casting based on the same 2D HPE as MeTRAbs.

\textbf{+ context.} Our method, improved by additional knowledge about this exact running scenario and ideal conditions. We base this method on the ground truth 2D joint locations and leverage the fact that running is cyclical. We automatically determine the frequency of the athlete's step and optimize the ray-casting portion of our approach to minimize the self-similarity between the same portion of a running cycle. This is a highly idealized scenario and is meant to showcase the minimal potential error that could be achieved under ideal conditions and with very specific domain knowledge.

\begin{figure}
    \centering
    \includegraphics[width=.45\textwidth]{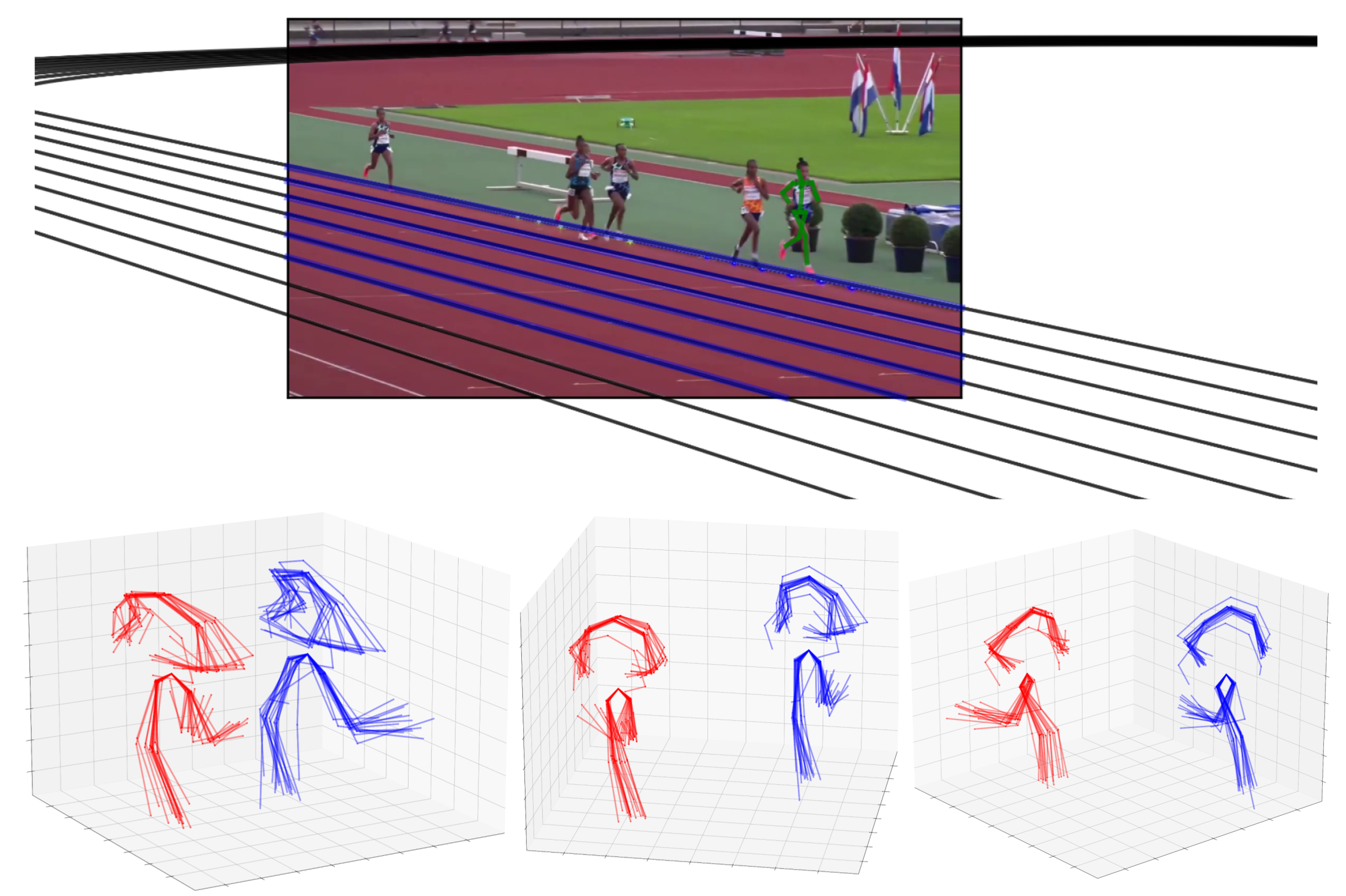}
    \caption{Application of our method to actual broadcast footage. Top: a scene with detected track markings (blue), projected stadium lanes (black), and 2D HPE (green). Bottom: Self-consistency of 3D pose estimation between similar frames in step-cycle for typical method (red) and our improvement (blue). There are notable differences in the estimation of the athlete's left shin.}
    \label{fig:anecdote}
\end{figure}

\subsection{Real-world footage}
The goal of this line of research is the application of monocular 3D HPE methods to a broad spectrum of broadcast footage in order to extract valid kinematic running data. We demonstrate the potential of our approach on anecdotal data from a world-class track~\&~field event. As there is no ground truth for this data, we demonstrate the qualitative differences between MeTRAbs+rotation and our method in \cref{fig:anecdote}. While increased self-similarity is not sufficient proof, that our method is correct, it is a necessary first step. The wider spread of the right knee of the red skeleton in \cref{fig:anecdote} illustrates the issue with non-specialized pose estimation approaches.

\subsection{Lens distortion}
Throughout the previous experiments, we assumed a pinhole camera without lens distortion. To test the impact that lens distortion has on our overall findings, we render a previously analyzed sequence with additional lens distortion.
We manually calibrate a single frame from a broadcasted running event (\cf \cref{fig:anecdote}), using the different lane and hurdle markers present in a frame. We identify 19 keypoints in the still image, for which the exact corresponding world coordinates are known. We perform an undistort calibration implemented in OpenCV to derive the parameters that we apply to our test scene. As expected, the highly zoomed-in viewpoint has only small lens distortion. 
We re-render our scene with a custom-built add-on to the camera in which we implement a simple lens distortion model, as used by Theiner and Ewerth~\cite{TVCalib22}. 
For the chosen scene, our method improves the 3D estimation error from 7.05 to 5.78cm. Performing the same evaluation on the distorted images results in an improvement from 9.88 to 8.08cm. While the distorted scene had a larger base error, there is still considerable improvement using our suggested method. 

\section{Discussion}
\label{sec:discussion}

In this work, we developed a method for combining 2D HPE and partial sports field registration to determine the 3D locations of joints and thereby the kinematics of athletes. On a synthetic dataset, we show that off-the-shelf pose estimation methods produce errors that would overshadow expected effect sizes in kinematic investigations. We suggest a specialized method with domain knowledge that could generate the required level of precision. Deploying our approach at scale and adapting it to further sports could help in collecting large datasets for the development of kinematically valid human motion representations.


Our work focuses on a scenario in which the camera is zoomed-in on the athlete at a level that kinematics are derivable from that view. In many sports (\eg, soccer), the field has an extent that does not allow to perform typical approaches of sports field registration in this case: not enough area of the field is visible. Specifically, if the entire 400m track was visible, the athlete would only occupy a few pixels in the image. Even in scenarios in which a curve of the field is visible, these must not always be pronounced enough to be used for a sports field registration task. We circumvent the problem of missing markers, by only taking into account a limited subset of straight lane markers. Even though we render the curved parts of the stadium in our figures, these were not used for the approximation of camera calibration.

It is tempting to leverage our synthetic dataset to train future iterations of typical monocular 3D HPE methods. After all, we mentioned that current methods carry inductive bias from their training sets, which are usually recorded in multi-camera settings and, therefore, have a reasonably small epipolar distance, which allows for only limited distances of the camera to the scene. Using our synthetic dataset, we could enhance previous training regimes with novel data. We discourage readers from doing so, as this would simply shift the issue of ambiguous 2D joint locations as illustrated in \cref{fig:perspective} to a new domain and the flattened 3D world information cannot be recovered. Instead, scene context and geometry should be considered whenever possible. The use-case of sports analysis lends itself nicely to this paradigm, as the dimensions of field markings are usually known.

Of course, our comparisons are highly unfair and skewed toward our exact scenario. That is the point we are trying to make in this investigation. If we want to improve the collection of data for a very narrow domain, special-purpose approaches should be developed, as off-the-shelf solutions for general use do not take advantage of the additional information that can be extracted from broadcast footage. 
By showcasing some shortcomings of the current state-of-the-art, we did not mean to criticize their approach, but rather to point out, that using the current benchmarks, training data, and paradigms, misses some opportunities, by using specialized domain knowledge. Since there is no ground truth and it is unlikely that there could ever be an actually recorded ground truth of the magnitude and variation of our data, we demonstrate qualitatively, what the developed improvements could mean in practice (\cf \cref{fig:anecdote}).

\subsection{Future work}
Optimizing the partial camera calibration and multiple options for pose estimation by ray-casting jointly could greatly improve the results of this work. For this, the single degree of freedom in the partial sports field registration could be parametrized in a single variable to allow for backpropagation through it. By using a differentiable projection framework, the remaining free parameters, like the limb lengths and cyclical consistency for a runner could be optimized jointly.

Our method can readily be adapted to other sports using existing camera calibration by sports field registration approaches~\cite{TVCalib22,chen2019sports}. Adapting our synthetic data generation process can be achieved by using sport-specific game-field assets and movements (\cf unreal marketplace, Mixamo).

Extracting valid 3D kinematics of athletes can be used, as initially motivated, to generate datasets that aim at large-scale understanding of human motion for a specific sport. In addition, providing coaches with valid kinematic data empowers them to make informed decisions on technique adjustments and training strategies, which could ultimately lead to performance enhancements by addressing biomechanical inefficiencies and optimizing athletes' movements. 

\subsection{Conclusion}
In this work, we presented an approach to improve the precision of 3D human pose estimation during the special use-case of running by injecting knowledge about the 3D geometry of the scene. To achieve this, our method generates a set of potential camera calibrations for each frame of a sequence and chooses a calibration for each frame to achieve location consistency of the camera over the entire sequence. Using this 3D geometry of the scene, the 3D location of an athlete is determined by calculating their sagittal plane and casting rays through it. Our method lowers the expected error for monocular 3D HPE in the special case of running to a kinematically valid point, enabling a more detailed and accurate analysis of human movement and performance at scale in the future.

{\small
\bibliographystyle{ieee_fullname}
\bibliography{references}
}

\end{document}